%% file: main.tex
\documentclass[10pt,twocolumn,letterpaper]{article}

\usepackage[pagenumbers]{cvpr} 

\input{preamble}
\definecolor{cvprblue}{rgb}{0.21,0.49,0.74}
\usepackage[pagebackref,breaklinks,colorlinks,allcolors=cvprblue, urlcolor=magenta]{hyperref}

\usepackage{multirow}
\usepackage{bm}
\usepackage{makecell}
\usepackage{cuted}
\usepackage{caption} 
\usepackage{adjustbox}
\usepackage[accsupp]{axessibility}  

\def\paperID{} 
\def\confName{CVPR}
\def\confYear{2026}

\title{Ground Reaction Inertial Poser: Physics-based Human Motion Capture \\ from Sparse IMUs and Insole Pressure Sensors}

\author{
Ryosuke Hori$^{1,2,3}$ \quad
Jyun-Ting Song$^{1}$ \quad
Zhengyi Luo$^{1}$ \quad
Jinkun Cao$^{1}$ \\
Soyong Shin$^{1}$ \quad
Hideo Saito$^{2,3}$ \quad
Kris Kitani$^{1}$ \\ 
$^{1}$Carnegie Mellon University \quad
$^{2}$Keio University \quad
$^{3}$Keio AI Research Center
\\
\small\tt \textcolor{magenta}{https://ryosukehori.github.io/grip-project/}
}

\begin{document}
\maketitle

\begin{strip}
  \centering
    \vspace{-13mm}
    \includegraphics[width=1.0\linewidth]{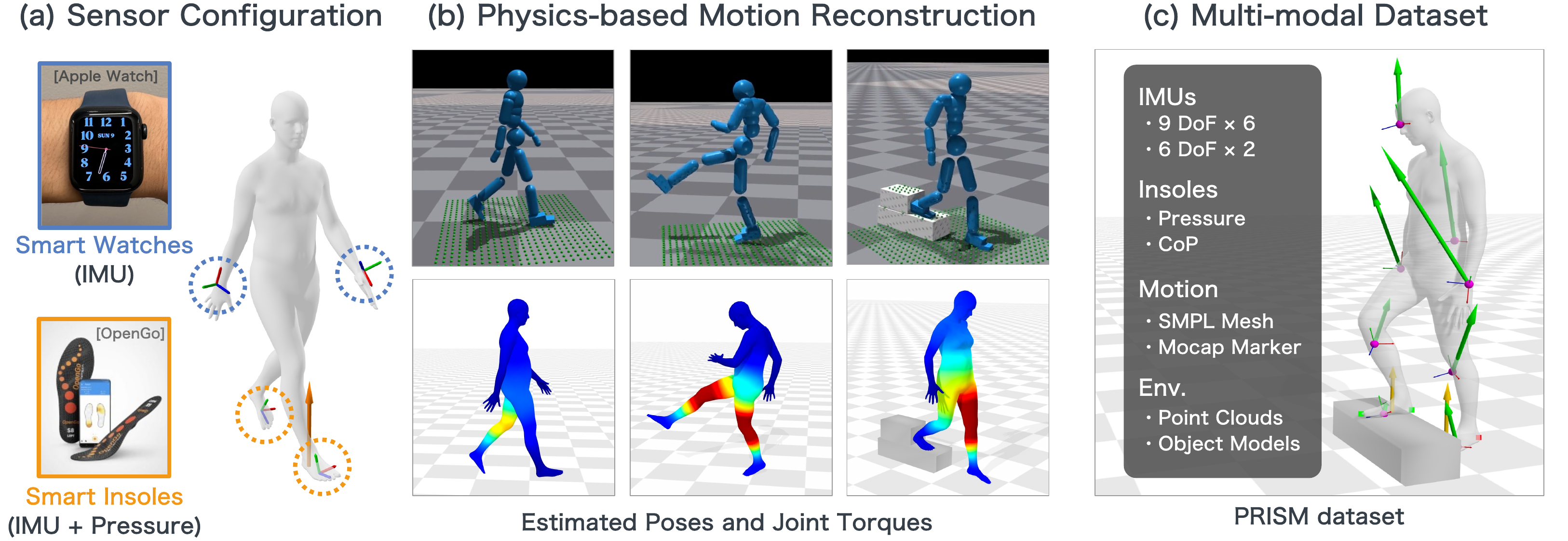}
    \captionof{figure}{
Overview of the proposed Ground Reaction Inertial Poser (GRIP).
(a) GRIP observes motion using four IMUs and foot pressure data from smartwatches and smart insoles.
(b) Full-body motion is reconstructed by driving a humanoid with joint torques in a physics simulator.
(c) The PRISM dataset offers multimodal measurements, including IMUs, foot pressure, motion data, and environmental data.
}
    \label{fig:overview}
\end{strip}

\input{sec/0_abstract}    
\input{sec/1_intro}
\input{sec/2_related_works}
\input{sec/3_method}

\input{sec/4_experiments}

\input{sec/5_coclusion}
\newpage
\paragraph{Acknowledgments.} 
This work was supported by JSPS Research Fellowships for Young Scientists DC1 and JSPS Overseas Challenge Program for Young Researchers.

{
    \small
    \bibliographystyle{ieeenat_fullname}
    \bibliography{main}
}

\input{sec/X_suppl}

\end{document}

%% file: sec/0_abstract.tex
\begin{abstract}
We propose Ground Reaction Inertial Poser (GRIP), a method that reconstructs physically plausible human motion using four wearable devices. Unlike conventional IMU-only approaches, GRIP combines IMU signals with foot pressure data to capture both body dynamics and ground interactions. Furthermore, rather than relying solely on kinematic estimation, GRIP uses a digital twin of a person, in the form of a synthetic humanoid in a physics simulator, to reconstruct realistic and physically plausible motion. At its core, GRIP consists of two modules: KinematicsNet, which estimates body poses and velocities from sensor data, and DynamicsNet, which controls the humanoid in the simulator using the residual between the KinematicsNet prediction and the simulated humanoid state. To enable robust training and fair evaluation, we introduce a large-scale dataset, Pressure and Inertial Sensing for Human Motion and Interaction (PRISM), that captures diverse human motions with synchronized IMUs and insole pressure sensors. Experimental results show that GRIP outperforms existing IMU-only and IMU–pressure fusion methods across all evaluated datasets, achieving higher global pose accuracy and improved physical consistency.

\end{abstract}

%% file: sec/1_intro.tex
\vspace{-3mm}
\section{Introduction}
\label{sec:intro}
Accurately capturing human motion in practical settings is crucial for many applications in robotics, biomechanics, and virtual reality. Reducing the number of sensors is particularly important for improving usability in everyday environments. In this work, we propose \textbf{Ground Reaction Inertial Poser (GRIP)}, a novel motion capture method that reconstructs physically consistent full-body motion using only four wearable devices attached to the wrists and feet (Fig.~\ref{fig:overview}-a). GRIP integrates kinematic cues from four IMUs with dynamic cues from foot pressure, and leverages physics simulation–based humanoid control through deep reinforcement learning, achieving accurate and physically plausible full-body motion estimation despite the minimal sensor configuration (Fig.~\ref{fig:overview}-b).

Conventional optical motion capture (MoCap) systems provide high accuracy, but require dedicated environments with multiple cameras, making them unsuitable for everyday or long-term use. Monocular RGB–based approaches have also been widely studied in recent years. However, third-person monocular methods~\cite{Cao_openpose, Fang_RMPE, Chen_monocular, Wang_survey, Tian_recovering} are constrained by the camera’s field of view, while egocentric methods~\cite{Hwang_MonoEye, Hu_EgoRenderer, Wang_EgoGlobal, Jiang_EgoSpan, Yi_EgoAllo, Guzov_HMD2} suffer from the difficulty of estimating the camera motion and the full-body pose due to self-occlusions. As an alternative to these vision-based approaches, wearable IMU-based methods, in which sensors are directly attached to the human body, have also been explored. Commercial wearable MoCap systems offer accurate motion estimation, but systems such as Xsens require up to 17 IMUs embedded in a spandex body suit, making them less suitable for daily use.

Recent studies address this limitation by estimating full-body pose using only a few IMUs~\cite{Huang_DIP, Yi_TransPose, Jiang_TIP, Mollyn_IMUPoser}. However, unlike methods that leverage VR-oriented head-mounted devices (HMDs) or hand trackers for pose estimation~\cite{vae_hmd, avatarposer, agrol, bodiffusion, avatarjlm, hmd_poser, sage, egoposer, flag, manikin, questsim, questenvsim}, IMU-only systems cannot directly measure absolute position. As a result, accumulated estimation errors lead to temporal drift and physically implausible artifacts, such as foot sliding, penetration, or floating. Although post-hoc physical optimization techniques~\cite{Yi_PIP, Yi_PNP, Xu_MobilePoser, Yi_GlobalPose} can mitigate these issues, they still depend on IMU-based estimation of body–environment contacts, making it difficult to accurately capture fine-grained physical interactions. To better model such interactions, recent studies incorporate insole-type pressure sensors equipped with IMUs~\cite{Wu_SolePoser, Gao_PressInPose, Ying_FoRM}, providing additional dynamic cues that reflect weight shifts through ground reaction forces, as well as reliable foot-contact cues that help suppress drift. Nevertheless, these approaches have several limitations, including reliance solely on foot sensors, limited upper-body motion reconstruction, lack of trajectory estimation, or positional drift without physical modeling.

To address these limitations, GRIP is designed to achieve high accuracy and physical plausibility with minimal IMUs and foot pressure sensors. It consists of two neural network modules. The first, KinematicsNet, incrementally estimates the kinematic state of motion, including joint positions, angles, and velocities, from IMU and foot pressure data. The second, DynamicsNet, controls a digital twin humanoid model within a physics simulator by using the sensor data, the estimated kinematic state, humanoid state, and environmental information as observations to reproduce physically consistent human motion.
By using the simulator, DynamicsNet naturally satisfies physical constraints such as gravity, friction, and ground reaction forces. This allows it to suppress non-physical artifacts, as well as unnatural foot-ground constraints seen in the physical optimization methods.

Validating such a physics-integrated framework requires real-world data that capture both human motion and its physical interactions with the environment.
To this end, we built a new public dataset, Pressure and Inertial Sensing for Human Motion and Interaction (\textbf{PRISM}), which integrates multi-modal data including IMU and insole pressure measurements, optical motion capture, and environmental models (Fig.~\ref{fig:overview}-c).
PRISM covers diverse human motions, spanning daily activities, sports, and human–object interactions, offering a comprehensive dataset for modeling physically consistent human motion and interaction.

In summary, our contributions are as follows: 
(1) We propose GRIP, a framework that achieves accurate full-body motion estimation using only four IMUs and insole pressure sensors. 
(2) We introduce a two-stage architecture, KinematicsNet and DynamicsNet, that integrates IMU and pressure data to reconstruct physically consistent human motion.
(3) We construct and publicly release a new multi-modal dataset, PRISM, which includes diverse actions and sensor modalities.

%% file: sec/2_related_works.tex
\section{Related Works}
\label{sec:related_works}

\subsection{Pose Estimation Using IMUs}

Full-body pose estimation using a small number of body-mounted IMUs has been widely studied. Early approaches demonstrated that accurate motion estimation is achievable even with sparse IMUs, through either optimization-based reconstruction~\cite{Marcard_SIP} or deep learning~\cite{Huang_DIP}. Subsequent studies have improved global localization accuracy and physical consistency by leveraging additional cues such as ground contact~\cite{Yi_TransPose}, physical constraints~\cite{Yi_PIP}, terrain estimation~\cite{Jiang_TIP}, non-inertial effects~\cite{Yi_PNP}, and contact reasoning for interactions with objects~\cite{Yi_GlobalPose}. Other works enhance estimation by integrating auxiliary sensing modalities, including UWB-based ranging~\cite{Armani2024UIP, Liu_UMotion}, barometric altitude information~\cite{Zhang_BaroPoser}, and egocentric vision~\cite{Guzov2021HPS, Xinyu2023EgoLocate, guzov24ireplica, lee2024mocapevery, Yin2024EgoHDM}. Recent research has further explored lightweight configurations using everyday devices~\cite{Mollyn_IMUPoser, Xu_MobilePoser, Zuo2024Loose, Hori_GIP}, as well as generative models that synthesize motions for body parts without IMU attachments~\cite{Wouwe2024Diffusion}.
However, reducing the number of sensors inherently makes it more difficult to estimate complex body motion and model body–environment interactions. These limitations suggest the need for additional complementary signals beyond IMUs.

\subsection{Pose Estimation Using Pressure Sensors}
Pressure sensors provide valuable information about human motion dynamics and human–environment interactions, and their integration has been shown to improve the physical plausibility of pose estimation~\cite{Luo_InteliCarpet, Chen_CAvatar, Tripathi_IPMAN, ren_MotionPRO}. However, these approaches rely on fixed pressure mats, limiting their applicability to constrained environments.
In parallel, portable insole-type pressure sensors have also been explored, enabling analyses of stability~\cite{Scott_Stability}, improvements in vision-based pose estimation~\cite{ Zhang_MMVP}, and reductions in foot-skating artifacts~\cite{Mourot_UnderPressure}. Moreover, several methods combine insole-mounted IMUs and pressure data to estimate root-relative poses~\cite{Wu_SolePoser, Gao_PressInPose} or to reconstruct both full-body posture and global trajectories using dual-modal inputs~\cite{Ying_FoRM}.
These IMU–pressure fusion approaches highlight the usefulness of foot–ground contact information; however, configurations restricted to the feet inherently struggle to capture full-body motion, and IMU measurements in local sensor coordinates make global trajectory estimation challenging. In addition, the existing methods lack mechanisms to enforce physical plausibility in reconstructed poses.
These limitations motivate our design of GRIP, which integrates sparse IMUs, insole pressure sensing, and physics simulation to achieve accurate and physically consistent full-body motion reconstruction with a minimal sensor configuration.

\subsection{Physics-based Humanoid Control}
Physics-based humanoid control has gained increasing attention in recent years. Prior work has demonstrated the ability to generate diverse motions from robust control policies and to generalize across varying environments~\cite{Luo_PHC, Luo_PULSE, tessler_MaskedMimic}. In the context of image-based pose estimation, several methods incorporate physical plausibility through reinforcement learning or physics-guided control to improve motion realism~\cite{Shimada_PhysCap, Shimada_NeuraPhysCap, Huang_NeuralMocon, luo2022embodied, Yuan2018, Yuan2019Ego, Luo2021DynamicsRegulatedKP, Zhang_Plug}. Additional advances have utilized knowledge distillation~\cite{Luo_SimXR} or positional signals from VR devices to enable real-time or adaptive physically consistent motion estimation~\cite{questenvsim}.
These approaches typically rely on absolute positional or postural information obtained from cameras or HMDs. In contrast, IMU-only settings lack such global references, and integrating IMU-derived velocities inevitably introduces drift, which directly affects the trajectories supplied to physics-based controllers. These limitations highlight the need for physics-based control frameworks that can operate without absolute positional inputs and instead utilize relative motion cues available from wearable sensors.

%% file: sec/3_method.tex
\begin{figure*}[t]
\centering
\includegraphics[width=0.99\textwidth]{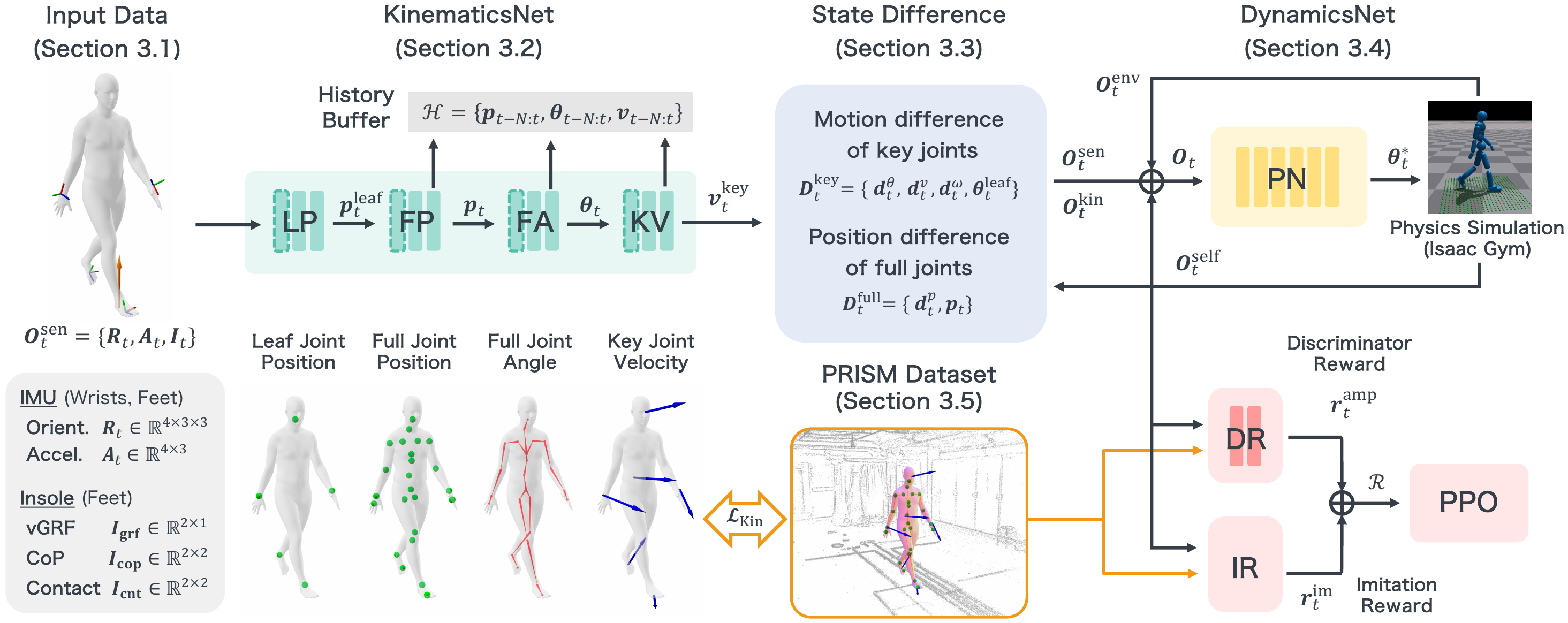}
\caption{
Overview of the GRIP framework.
\textbf{Input Data} (Sec.~\ref{subsec:input_data}) consists of IMU and insole measurements.
\textbf{KinematicsNet} (Sec.~\ref{subsec:kin_net}) estimates kinematic states, and the \textbf{State Difference} (Sec.~\ref{subsec:state_diff}) compares them with the simulated humanoid.
\textbf{DynamicsNet} (Sec.~\ref{subsec:dyn_net}) drives the humanoid through physics simulation-based control.
The \textbf{PRISM dataset} (Sec.~\ref{subsec:prism}) provides diverse multi-modal data.
}
\label{fig:network}
\end{figure*}

\section{Method}
\label{sec:method}
Our goal is to reproduce diverse real-world human motions on a torque-driven humanoid model in a physics simulator using a minimal sensor setup. As shown in Fig.~\ref{fig:network}, GRIP takes four IMU signals, two on the wrists and two embedded in the insoles, along with the foot pressure data. The architecture follows an observer–controller structure: KinematicsNet estimates motion states from sensor data, and DynamicsNet uses these estimates to control the humanoid model. We introduce an intermediate representation, State Difference, which captures the discrepancy between the estimated motion state and the simulated humanoid state, allowing DynamicsNet to utilize the kinematic observations. The following sections describe the input data (Sec.~\ref{subsec:input_data}), KinematicsNet (Sec.~\ref{subsec:kin_net}), State Difference (Sec.~\ref{subsec:state_diff}), DynamicsNet (Sec.~\ref{subsec:dyn_net}), and the PRISM dataset (Sec.~\ref{subsec:prism}).

\subsection{Input Data}
\label{subsec:input_data}
The GRIP framework uses acceleration and orientation measurements from four IMUs. All IMU accelerations and orientations are transformed into the global coordinate frame. For a sequence of length $T$, the acceleration and orientation at frame $t$ are represented as $\bm{A}_t \in \mathbb{R}^{4\times3}$ and $\bm{R}_t \in \mathbb{R}^{4\times3\times3}$.
The insole sensors contain 16 pressure cells per foot and provide raw pressure values, vertical ground reaction forces (GRF), and the center of pressure (CoP). To ensure compatibility across different insole devices, we define a simplified input representation. Specifically, the vertical GRF is represented as $\bm{I}_{\text{grf}} \in \mathbb{R}^{2\times1}$, the CoP positions (2D coordinates centered at the midfoot) as $\bm{I}_{\text{cop}} \in \mathbb{R}^{2\times2}$, and the binary contact labels for the forefoot and rearfoot regions as $\bm{I}_{\text{cnt}} \in \mathbb{R}^{2\times2}$. The contact labels are obtained by dividing the foot sole into front and rear regions and assigning a contact label when the summed pressure in each region exceeds a threshold.
These features are concatenated to form the simplified input representation $\bm{I}_t = [\bm{I}_{\text{grf}}, \bm{I}_{\text{cop}},\bm{I}_{\text{cnt}}] \in \mathbb{R}^{2\times5}$. Finally, all sensor observations at frame $t$ are summarized as $\bm{O}^{\text{sen}}_t = \{\bm{R}_t,\ \bm{A}_t,\ \bm{I}_t\}$.

\subsection{KinematicsNet}
\label{subsec:kin_net}
KinematicsNet progressively estimates kinematic information from the input sensor data. The network produces four outputs: leaf-joint positions (LP), full-joint positions (FP), full-body joint angles (FA), and key-joint velocities (KV). First, taking the root joint (pelvis) as the coordinate origin, the network predicts the leaf-joint positions (LP) of the wrists, feet, and head. These predictions, together with the sensor data, are then used to reconstruct full-body joint positions (FP). While this progressive estimation follows prior works~\cite{Yi_TransPose, Yi_PIP, Yi_PNP, Yi_GlobalPose}, our method differs in that we operate directly on globally oriented IMU measurements. Because no IMU is attached to the root joint, the root’s global rotation cannot be separated from these measurements. As a result, the network outputs are represented in a root-centered frame but still retain the global rotation of the body.
Next, using the sensor data and the estimated joint positions, the network predicts the full-body joint angles (FA) and the global key-joint velocities (KV). The resulting full-joint positions, joint angles, and key-joint velocities are used to compute the State Difference (Sec.~\ref{subsec:state_diff}), and are also stored in a History Buffer $\mathcal{H}$ to provide reset references when the humanoid falls in the physics simulator (Sec.~\ref{subsec:dyn_net}).

The network employs a unidirectional LSTM for each stage (LP, FP, FA, KV), estimating outputs frame by frame while modeling temporal dependencies. At each timestep, the output is defined as leaf-joint position $\bm{p}_t^{\text{leaf}} \in \mathbb{R}^{5\times 3}$, full-joint position $\bm{p}_t \in \mathbb{R}^{24\times 3}$, full-body joint angle $\bm{\theta}_t \in \mathbb{R}^{24\times 6}$, and key-joint velocity $\bm{v}_t^{\text{key}} \in \mathbb{R}^{6\times 3}$. The 24 joints correspond to the SMPL model~\cite{smpl}, the five leaf joints correspond to the wrists, feet, and head, and the six key joints include the leaf joints and the root joint. The joint angles are represented using the continuous 6D rotation representation~\cite{Zhou_Continuity} for stable learning. KinematicsNet is trained using mean squared error (MSE) losses applied to each of its submodules. The total loss is defined as: $
\mathcal{L}_{\text{Kin}} =
\|\bm{p}^{\text{leaf}} - \hat{\bm{p}}^{\text{leaf}}\|^{2}
+ \|\bm{p} - \hat{\bm{p}}\|^{2} 
+ \|\bm{\theta} - \hat{\bm{\theta}}\|^{2}
+ \|\bm{v}^{\text{key}} - \hat{\bm{v}}^{\text{key}}\|^{2},
$
where the hat notation indicates ground-truth values.

\subsection{State Difference}
\label{subsec:state_diff}
In IMU-based motion estimation, integrating estimated velocities to obtain global trajectories causes drift accumulation, which would directly propagate to humanoid tracking. To avoid this, our framework does not use integrated global positions; instead, it represents translational motion using the estimated global velocities of key joints and expresses posture using root-relative joint positions. To unify these cues and capture the discrepancy between the estimated and simulated humanoid states, we introduce an intermediate representation called State Difference.

The State Difference comprises (1) motion differences at the key joints and (2) position differences for all joints, both transformed into the humanoid’s heading-aligned coordinate frame.
The first component, $\bm{D}^{\text{key}}_t = \{\bm{d}^\theta_t, \bm{d}^v_t, \bm{d}^\omega_t, \bm{\theta}^{\text{leaf}}_t\}$,  
consists of the following terms:  
the rotational differences $\bm{d}^\theta_t \in \mathbb{R}^{4\times6}$,  
angular velocity differences $\bm{d}^\omega_t \in \mathbb{R}^{4\times3}$,  
and orientations $\bm{\theta}^{\text{leaf}}_t \in \mathbb{R}^{4\times6}$ computed at the four leaf joints corresponding to the IMUs (wrists and feet);  
and the linear velocity differences $\bm{d}^v_t \in \mathbb{R}^{6\times3}$  
computed at six key joints (wrists, feet, head, and pelvis) based on the outputs of KinematicsNet.
The second component, $\bm{D}^{\text{full}}_t = \{\bm{d}^p_t, \bm{p}_t\}$,  
captures the position differences of all joints, expressed in the root-aligned coordinate frame.  
Specifically, the position difference $\bm{d}^p_t \in \mathbb{R}^{24\times3}$ 
are derived from the full-body joint positions $\bm{p}_t \in \mathbb{R}^{24\times3}$  estimated by KinematicsNet and the corresponding simulated humanoid joint positions.  
Finally, $\bm{D}^{\text{key}}_t$ and $\bm{D}^{\text{full}}_t$ are concatenated to form the State Difference vector  
$\bm{D}_t = [\bm{D}^{\text{key}}_t, \bm{D}^{\text{full}}_t]$.

\subsection{DynamicsNet}
\label{subsec:dyn_net}

DynamicsNet controls the humanoid model within the physics simulator based on several observations.  
The overall framework is formulated as a Markov Decision Process defined by the tuple 
$\mathcal{M} = \langle \mathcal{S}, \mathcal{A}, \mathcal{T}, \mathcal{R}, \gamma \rangle$,  
where $\mathcal{S}$, $\mathcal{A}$, $\mathcal{T}$, $\mathcal{R}$, and $\gamma$ denote the state space, action space, transition dynamics, reward function, and discount factor, respectively.  
At each timestep $t$, the policy $\pi(\bm{a}_t|\bm{s}_t)$ selects an action $\bm{a}_t \in \mathcal{A}$ given the current state $\bm{s}_t \in \mathcal{S}$,  
and the simulator updates the next state $\bm{s}_{t+1}$ according to $\mathcal{T}$.  
The policy is trained to maximize the expected discounted reward using the proximal policy optimization (PPO) algorithm~\cite{Schulman2017PPO}.

\vspace{0.5\baselineskip}
\noindent\textbf{Observations.}
The observation vector at each timestep is defined as
$\bm{O}_t = \{\bm{O}_t^{\text{sen}}, \bm{O}_t^{\text{kin}}, \bm{O}_t^{\text{self}}, \bm{O}_t^{\text{env}}\}$.
The sensor observation $\bm{O}_t^{\text{sen}} = \{\bm{R}_t, \bm{A}_t, \bm{I}_t\}$ consists of the same inputs used in KinematicsNet.
These sensor-level features allow the policy to directly reference the original sensory inputs in addition to the estimated kinematic states.
The kinematic observation $\bm{O}_t^{\text{kin}}$ corresponds to the State Difference $\bm{D}_t$ introduced in the previous section, representing the discrepancies between the estimated motion and simulated humanoid states.
The self observation $\bm{O}_t^{\text{self}} = \{\bm{p}^\text{sim}_t, \bm{\theta}^\text{sim}_t, \dot{\bm{p}}^\text{sim}_t, \dot{\bm{\theta}}^\text{sim}_t\}$ includes the positions and angles of all joints, along with their temporal derivatives (velocities and angular velocities).
Finally, the environmental observation $\bm{O}_t^{\text{env}}$ represents the height map of the environment, sampled as a $25\times25$ grid centered around the humanoid within a square area of $1.5 \text{m} \times 1.5 \text{m}$.
This map provides spatial context, enabling the policy to adapt to non-flat environments.

\vspace{0.5\baselineskip}
\noindent\textbf{Policy and Reward.}
The policy network (PN) is implemented as a multilayer perceptron (MLP) that maps the observation vector $\bm{O}_t$  
to target joint angles $\bm{\theta}^*_t$, which serve as the action $\bm{a}_t$.  
These target angles are converted into joint torques $\bm{\tau}_t$ using proportional-derivative (PD) control:
$\bm{\tau}_t = k_p(\bm{\theta}^*_t - \bm{\theta}_t) - k_d\dot{\bm{\theta}}_t,
$
where $k_p$ and $k_d$ denote proportional and derivative gains, respectively.
The reward function follows the Perpetual Humanoid Control (PHC) framework~\cite{Luo_PHC} and consists of three components: an Adversarial Motion Prior (AMP)~\cite{Peng_AMP}, an imitation reward, and an energy penalty~\cite{Fu2022DeepWholeBodyControl}.
The discriminator reward (DR) from AMP encourages realistic motion by training a discriminator to distinguish generated motions from real human movements.
The imitation reward (IR) measures consistency between the simulated and reference motion obtained from the training dataset using differences in joint positions, rotations, and velocities.
The energy penalty regularizes excessive joint torques, promoting natural and stable motion.
Further details are provided in the supplementary material.

\vspace{0.5\baselineskip}
\noindent\textbf{Fall Recovery Mechanism.}
In our framework, the humanoid is controlled under strong physical constraints without introducing any auxiliary forces such as residual forces.  
As a result, during complex or abrupt motion transitions, the humanoid may occasionally lose balance and fall, leading to unstable simulation behavior.  
To address this issue, we introduce a fall recovery mechanism based on a history buffer to ensure stable inference during deployment.

The mechanism maintains a history buffer $\mathcal{H} = \{\bm{p}_{t-N:t}, \bm{\theta}_{t-N:t}, \bm{v}_{t-N:t}\}$ containing the past $N$ frames of outputs from KinematicsNet.  
At each timestep $t$, a fall is detected when the root joint height $\bm{p}^{\text{root}}_{z,t}$ falls below the threshold $\tau_z$ and the discriminator probability $\rho_t$ from the AMP module is lower than $\tau_\rho$.
Once a fall is detected, we compute the kinematic root displacement
$\Delta \bm{p}^{\text{kin, root}}_{t-N:t}$
by integrating the root velocities estimated by KinematicsNet and stored in the buffer.
The simulated root position is then reinitialized as
$\bm{p}^{\text{sim, root}}_t = \bm{p}^{\text{sim, root}}_{t-N} + \Delta \bm{p}^{\text{kin, root}}_{t-N:t}$.  
At the same time, the joint angles $\bm{\theta}_t$ estimated by the Full-joint Angle (FA) module of KinematicsNet are used to reset the humanoid states, after which the physical simulation is resumed.  
In addition, during the falling phase, the corresponding segment of motion output is replaced by the buffered KinematicsNet predictions $\bm{p}_{t-N:t}$, $\bm{\theta}_{t-N:t}$, and $\bm{v}_{t-N:t}$ to maintain kinematic continuity and prevent corrupted physical transitions.  
This allows the humanoid to continue its motion smoothly and consistently even after a fall, without requiring external correction forces.

\subsection{PRISM Dataset}
\label{subsec:prism}
To enable robust training and fair evaluation of GRIP, we require data captured under diverse conditions with multiple IMUs, insole pressure sensors, and high-fidelity motion labels. However, existing pressure-based human motion datasets~\cite{Scott_Stability, Mourot_UnderPressure, Han_GroundLink, Zhang_MMVP, Ying_FoRM} are limited in sensor modalities, motion diversity, or pose annotation accuracy, making them insufficient for training and evaluating physically consistent pose estimation from sparse sensors.

To address these limitations, we constructed a new multimodal dataset, Pressure and Inertial Sensing for Human Motion and Interaction (PRISM), that covers a broad set of motions, such as daily activities (e.g., walking, jogging), slow movements (e.g., stretching, squats), fast sports actions (e.g., golf, baseball, soccer), and interaction behaviors with objects (e.g., stepping onto or sitting on objects). IMUs, insole sensors, and optical MoCap were recorded at 100 Hz. IMU-equipped watches and insoles were attached to the wrists and feet, while additional IMUs were placed on the knees, pelvis, and head. SMPL pose labels were obtained by fitting the mesh model to the MoCap markers with an existing method~\cite{Loper_Mosh}. PRISM consists of 1,275 ten-second sequences collected from six subjects (four male and two female), totaling about 3.5 hours of synchronized multimodal motion data. Details on sensor specifications, calibration, synchronization, and comparisons with existing datasets are provided in the supplementary material.

%% file: sec/4_experiments.tex
\begin{table*}[t]
\centering
\caption{Quantitative comparison of our method with baseline methods across the three datasets. Lower values indicate better performance for all metrics. Bold numbers denote the best performance for each metric, and underlined numbers denote the second best.}
\adjustbox{max width=\textwidth}{
\begin{tabular}{clcccccccc}
\toprule
Dataset & Method &
\makecell[c]{MPJPE\\{[mm]} (↓)} &
\makecell[c]{PEL-MPJPE\\{[mm]} (↓)} &
\makecell[c]{PA-MPJPE\\{[mm]} (↓)} &
\makecell[c]{MPJRE\\{[deg]} (↓)} &
\makecell[c]{Acc\\{[m/s$^2$]} (↓)} &
\makecell[c]{FS\\{[m/s]} (↓)} &
\makecell[c]{FP\\{[mm]} (↓)} &
\makecell[c]{vGRF\\{[N]} (↓)} \\
\midrule

\multirow{6}{*}{PRISM} 
& PIP          & 248.59 & 85.48 & \underline{33.35} & 17.08 & 6.68 & \underline{0.20} & 10.71 & \textbf{246.85} \\
& GlobalPose   & \underline{198.30} & \textbf{43.50} & \textbf{31.29} & \textbf{12.01} & 7.72 & 0.22 & \underline{9.72} & 299.22 \\
& MobilePoser  & 267.45 & 72.76 & 55.69 & 17.99 & \textbf{6.20} & \textbf{0.19} & 9.97 & \underline{248.54} \\
& FoRM         & 199.60 & 87.34 & 63.75 & 20.23 & 9.67 & 0.36 & 15.64 & - \\
& SolePoser    & - & - & 82.75 & - & 9.68 & - & - & - \\
& \textbf{GRIP (Ours)}& \textbf{182.44} & \underline{63.85} & 46.47 & \underline{13.89} & \underline{7.30} & 0.21 & \textbf{5.77} & 258.40 \\
\midrule

\multirow{6}{*}{\makecell{Under\\Pressure}} 
& PIP          & 523.65 & \underline{29.89} & \underline{21.08} & 8.35 & \textbf{12.59} & \underline{0.32} & 1.43 & \underline{265.62} \\
& GlobalPose   & \underline{301.12} & \textbf{21.49} & \textbf{17.41} & \textbf{7.40} & 16.65 & 0.32 & 3.31 & 287.12 \\
& MobilePoser  & 626.62 & 44.28 & 33.74 & 11.73 & \underline{12.62} & 0.35 & \underline{1.27} & \textbf{244.78} \\
& FoRM         & 553.19 & 59.29 & 32.60 & 18.50 & 17.34 & 0.57 & 21.52 & - \\
& SolePoser    & - & - & 34.52 & - & 24.27 & - & - & - \\
& \textbf{GRIP (Ours)}& \textbf{218.09} & 37.27 & 27.16 & \underline{7.64} & 13.22 & \textbf{0.31} & \textbf{0.00} & 278.27 \\
\midrule

\multirow{6}{*}{\makecell{PSU-\\TMM100}} 
& PIP          & 182.14 & 87.56 & 61.62 & 21.12 & 1.62 & \textbf{0.09} & 5.84 & 367.87 \\
& GlobalPose   & 175.96 & \textbf{63.05} & \textbf{50.28} & \underline{18.71} & 2.09 & 0.14 & \underline{2.95} & \underline{340.11} \\
& MobilePoser  & 210.66 & 112.35 & 85.46 & 28.10 & 2.37 & \underline{0.10} & 5.28 & 358.19 \\
& FoRM         & \underline{126.60} & 98.02 & 82.45 & 25.19 & \underline{1.60} & 0.13 & 4.51 & - \\
& SolePoser    & - & - & 97.11 & - & \textbf{1.36} & - & - & - \\
& \textbf{GRIP (Ours)}& \textbf{118.60} & \underline{70.32} & \underline{55.72} & \textbf{16.72} & 4.31 & 0.11 & \textbf{0.73} & \textbf{316.06} \\
\bottomrule
\end{tabular}
}
\label{tab:comparison_results}
\end{table*}

\section{Experiments}
\label{sec:experimets}

\subsection{Implementation Details}
\noindent\textbf{Networks and Training.}
Our framework is trained in two stages: supervised learning of KinematicsNet and reinforcement learning of DynamicsNet within a physics simulatior.
KinematicsNet is first trained by optimizing its four submodules independently, followed by joint fine-tuning of the entire network. For training DynamicsNet, the weights of KinematicsNet are frozen, and the policy is optimized using PPO with the physics simulator. The fall recovery mechanism is used only during inference and not during training.
Details of the network architectures, hyperparameters, rewards, simulation settings, and computational environment are provided in the supplementary material.

\begin{table}[t]
\centering
\caption{Comparison of input modalities, output representations, and physical modeling across baseline methods.}
\label{tab:comparison_overview}
\resizebox{\linewidth}{!}{
\begin{tabular}{l|cc|ccc|cc}
\hline
\multirow{2}{*}{Method} & \multicolumn{2}{c|}{Input Data} & \multicolumn{3}{c|}{Output Data} & \multicolumn{2}{c}{Physics} \\ 
 & IMUs & Press. & Joints & Mesh & Transl. & Optim. & Sim. \\ \hline
PIP~\cite{Yi_PIP} & 6 &  & \checkmark & \checkmark & \checkmark & \checkmark &  \\
GlobalPose~\cite{Yi_GlobalPose} & 6 &  & \checkmark & \checkmark & \checkmark & \checkmark &  \\
MobilePoser~\cite{Xu_MobilePoser} & 3 &  & \checkmark & \checkmark & \checkmark & \checkmark &  \\
FoRM~\cite{Ying_FoRM} & 2 & \checkmark & \checkmark & \checkmark & \checkmark &  &  \\
SolePoser~\cite{Wu_SolePoser} & 2 & \checkmark & \checkmark &  &  &  &  \\
\textbf{GRIP (Ours)} & 4 & \checkmark & \checkmark & \checkmark & \checkmark &  & \checkmark \\ \hline
\end{tabular}}
\end{table}

\vspace{0.5\baselineskip}
\noindent\textbf{Datasets.}
We use three datasets with distinct motion characteristics: \textbf{PRISM}, \textbf{UnderPressure}~\cite{Mourot_UnderPressure}, and \textbf{PSU-TMM100}~\cite{Scott_Stability}.
PRISM contains diverse real-world motions, including daily activities, sports actions, and human–object interactions, captured with real IMUs and insole pressure sensors.
UnderPressure primarily consists of locomotion-centric actions such as walking, jogging, and skipping; although some sequences include interactions such as stair climbing or stepping onto objects, these were excluded in our experiments because object models are not provided.
PSU-TMM100 focuses on Tai Chi movements, which involve slow, highly controlled motions with rich balance shifts.
Since PRISM includes only six subjects, we randomly split all sequences into an 8:2 ratio for training and testing.
For UnderPressure and PSU-TMM100, which each contain ten subjects, eight were used for training and two for testing.
Although both datasets provide real insole measurements, they do not contain IMU recordings; therefore, following prior work~\cite{Yi_TransPose}, we synthesized IMU signals by computing finite differences of the vertex orientations at IMU attachment locations on the SMPL mesh.
All datasets were evaluated in 500-frame (5-second) segments.

\begin{figure*}[t]
\centering
\includegraphics[width=1.0\textwidth]{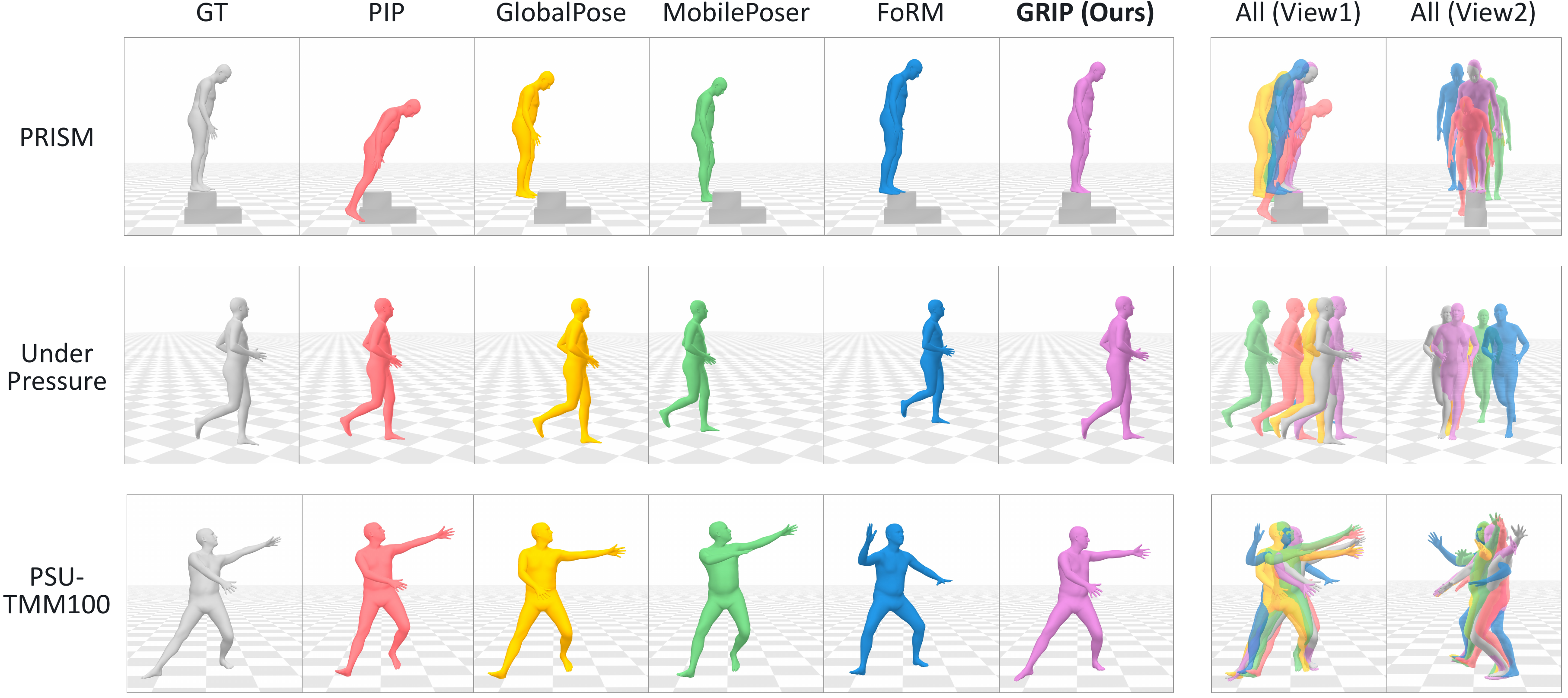}
\caption{Qualitative comparison of pose estimation results across the three datasets. Our method accurately reconstructs foot placement on objects (PRISM), exhibits less position drift (UnderPressure), and captures slow weight-shifting motions (PSU-TMM100).}
\label{fig:compare1}
\vspace{-2mm}
\end{figure*}

\vspace{0.5\baselineskip}
\noindent\textbf{Metrics.}
We evaluate both the proposed and baseline methods in terms of kinematic accuracy and physical plausibility.
For kinematic accuracy, we compute Mean Per Joint Position Error (\textbf{MPJPE}); Pelvis-Aligned MPJPE (\textbf{PEL-MPJPE}), which removes global translation; and Procrustes-Aligned MPJPE (\textbf{PA-MPJPE}), which further removes global scale and rotation.
We also evaluate joint rotation accuracy using Mean Per Joint Rotation Error (\textbf{MPJRE}), and temporal smoothness using Acceleration Error (\textbf{Acc}).
For physical plausibility, we measure Foot Sliding (\textbf{FS}) during contact and Foot Penetration Depth (\textbf{FP}) below the ground or object surface.
To assess dynamic consistency, we compute the Vertical Ground Reaction Force Error (\textbf{vGRF}), applied to baselines with physical optimization and to our method.
Metric units are reported in Table~\ref{tab:comparison_results}.

\begin{figure}[t]
\centering
\includegraphics[width=0.48\textwidth]{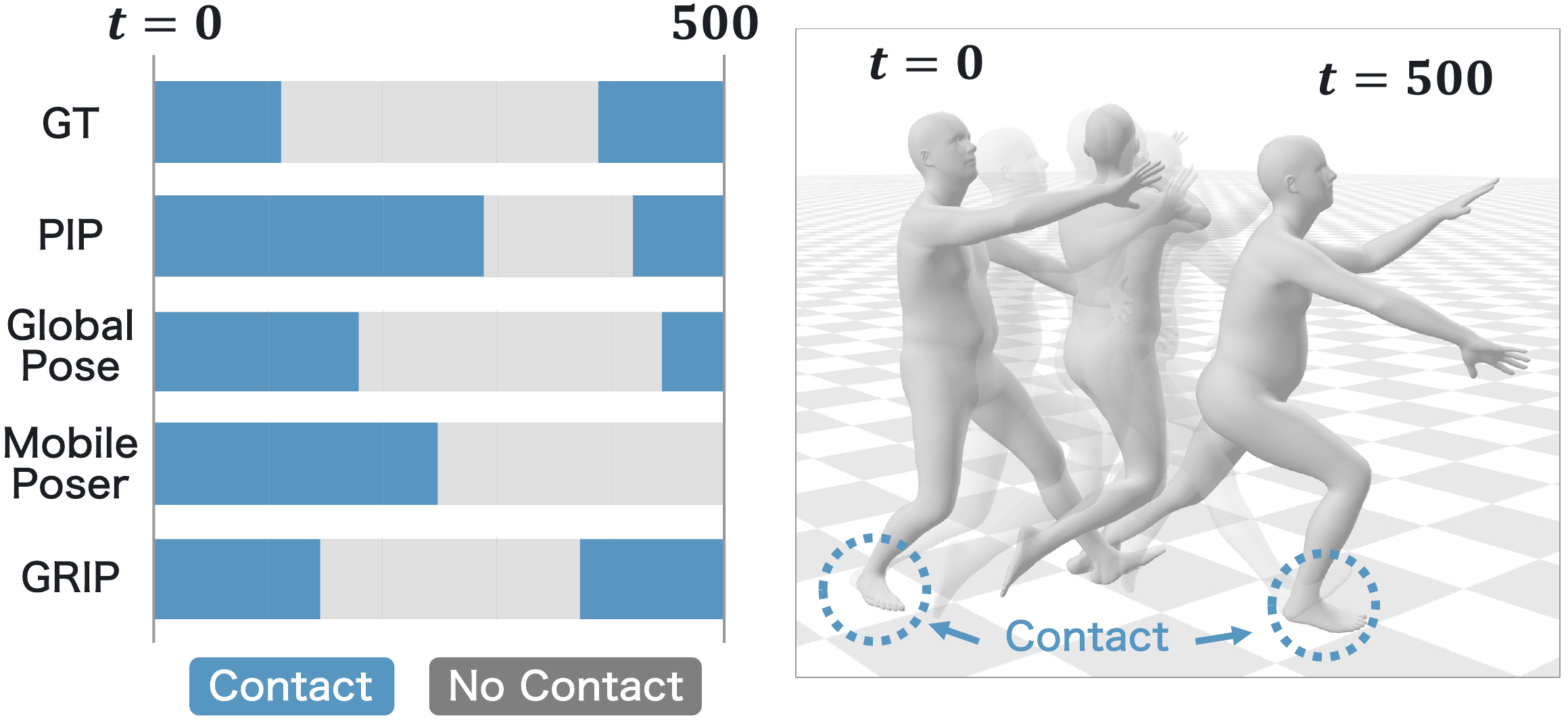}
\caption{Comparison of foot contact timing. Right-foot contact labels during low-speed motions in PSU-TMM100, computed from the estimated GRF of the physics-based methods.}
\label{fig:contact}
\vspace{-2mm}
\end{figure}

\begin{figure}[t]
\centering
\includegraphics[width=0.47\textwidth]{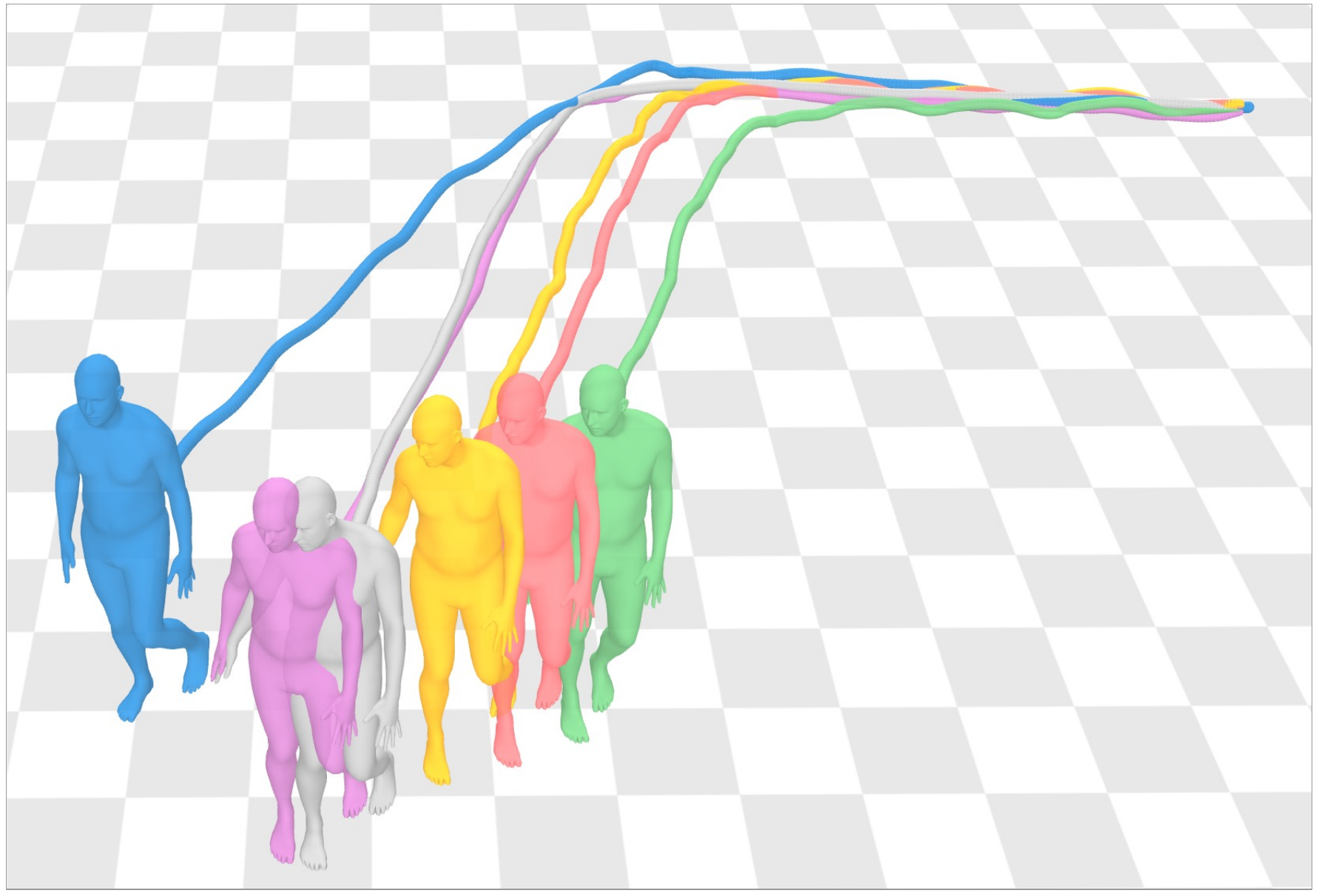}
\caption{Qualitative comparison of estimated poses and root trajectories for a walking sequence from the UnderPressure dataset. Colors correspond to the same methods shown in Fig.~\ref{fig:compare1}.}
\label{fig:trajectory}
\vspace{-2mm}
\end{figure}

\subsection{Comparison}
\label{subsec:comparison}
We compare our method with baseline methods that estimate full-body motion using IMUs and insole pressure sensors.
Table~\ref{tab:comparison_overview} summarizes the input modalities, output representations, and whether each method incorporates physical modeling.
Quantitative results are presented in Table~\ref{tab:comparison_results}, and qualitative results are shown in Figs.~\ref{fig:compare1},~\ref{fig:contact} and~\ref{fig:trajectory}.

For global pose accuracy (MPJPE), our method achieves the best performance across all datasets, PRISM (diverse motions), UnderPressure (large displacements), and PSU-TMM100 (slow motions with weight shifts).
The foot-sliding metric (FS) shows that physics-based approaches, including post-hoc optimization methods (PIP, GlobalPose, MobilePoser) and our simulation-based method, suppress foot sliding more effectively than the purely kinematic FoRM. However, optimization-based methods may over-apply contact constraints when estimation errors occur, temporarily locking the foot to the ground and resulting in trajectory errors (Fig.~\ref{fig:contact}). In contrast, our method simulates natural foot–ground interactions, enabling physically plausible and accurate trajectories (Fig.~\ref{fig:trajectory}).
For pose metrics without global translation (PEL-MPJPE, PA-MPJPE, MPJRE), GlobalPose achieves the highest accuracy due to using six IMUs and a correction mechanism based on a pelvis-mounted IMU that learns gravity–pose relationships. PIP and our four-IMU method achieve comparable accuracy, as the use of pressure information effectively captures body dynamics and compensates for the absence of pelvis and head IMUs. SolePoser and FoRM, which rely only on insole-mounted IMUs and pressure signals, are inherently limited in reconstructing accurate full-body motion.

For the foot penetration metric (FP), our method achieves the lowest values across all datasets.
In flat-ground datasets such as UnderPressure and PSU-TMM100, the three physics-optimization methods exhibit penetration only when contact detection fails, and our method also shows only minor penetration in limited situations, such as during state replacement during a fall. All four physics-based methods fundamentally prevent penetration through physical constraints.
In contrast, the purely kinematic method FoRM is more prone to penetration caused by root-position estimation errors, particularly in highly dynamic sequences such as PRISM and UnderPressure.
In the PRISM dataset, which contains objects, our method accurately handles foot–object interactions through a physics-based object model and achieves higher accuracy than MobilePoser and PIP, which assume flat-ground contact, and GlobalPose, which optimizes height and support states but may still suffer from imperfect contact detection.
However, in our method, errors may still occur in cases such as stumbling or stepping off elevated surfaces, where the estimated foot position temporarily falls below the object surface.

For vGRF error, PIP and MobilePoser achieve the highest accuracy in PRISM and UnderPressure, while our method performs best in PSU-TMM100.
The three optimization-based methods adopt floating-base models and stabilize physics-based control by applying residual forces.
In contrast, our method employs a humanoid model without floating bases or residual forces, which can lead to foot-pressure patterns that differ from those of real humans during balance loss or recovery in high-momentum motions.
However, in slower motions such as those in PSU-TMM100, the other methods tend to miss fine-scale foot contact events and consequently produce larger ground-reaction and position errors.
Our method maintains higher accuracy in these cases by combining natural foot–ground interaction through physics simulation with pressure-driven dynamic reasoning. Further details and extended results can be found in the Supplementary Material.

\subsection{Evaluation}
\noindent\textbf{Sensor Configuration Analysis.}
This experiment examines how different sensor configurations affect estimation accuracy and motion stability using the PRISM dataset.
We evaluated ten conditions: five IMU configurations (2–6 IMUs), each tested with and without insole pressure sensors.
IMUs were incrementally attached to both feet, both wrists (left only for the 3-IMU setup), the pelvis, and the head.
MPJPE and PA-MPJPE were used for kinematic accuracy, and the Success Rate measured the proportion of sequences completed without falling.
The quantitative results in Table~\ref{tab:sensor_config_results} show that accuracy improves with more IMUs and further increases when foot pressure information is added.
These findings indicate that increasing the number of IMUs enhances motion estimation accuracy, while foot pressure serves as a complementary modality that consistently improves both pose estimation and stability.

\begin{table}[h]
\centering
\caption{Comparison across different sensor configurations.}
\begin{tabular}{cc|ccc}
\toprule
\#IMU & Press.
& \makecell{MPJPE \\ \textnormal{[mm]} ($\downarrow$)}
& \makecell{PA-MPJPE \\ \textnormal{[mm]} ($\downarrow$)}
& \makecell{Succ. Rate \\ \textnormal{[\%]} ($\uparrow$)} \\
\hline
2  &  & 262.01 & 84.47 & 90.35 \\
3  &  & 209.27 & 57.30 & 88.98 \\
4  &  & 194.48 & 49.97 & 88.58 \\
5  &  & 186.95 & 45.68 & 86.61 \\
6  &  & \underline{160.76} & \underline{41.44} & 87.01 \\
\hline
2  & \checkmark & 247.12 & 82.36 & \underline{93.90} \\
3  & \checkmark & 199.60 & 56.80 & 91.14 \\
4  & \checkmark & 182.44 & 46.47 & \textbf{94.49} \\
5  & \checkmark & 164.25 & 42.82 & 90.94 \\
6  & \checkmark & \textbf{143.06} & \textbf{39.13} & 92.32 \\
\bottomrule
\end{tabular}
\label{tab:sensor_config_results}
\end{table}
\vspace{-2mm}

\vspace{0.5\baselineskip}
\noindent\textbf{Ablation Study on Observation Design.}  
This experiment examines how the kinematic information estimated by KinematicsNet, used within the State Difference, affects the control performance of DynamicsNet using the PRISM dataset. Sensor observations $\bm{O}^{\text{sen}}_t$, self observations $\bm{O}^{\text{self}}_t$, and environment observations $\bm{O}^{\text{env}}_t$ are used across all settings, while the kinematic observation $\bm{O}^{\text{kin}}_t$ is gradually expanded. Specifically, we evaluate four configurations:
(1) $\bm{D}(O, A)$ using orientation and acceleration differences from IMUs;
(2) $\bm{D}(O, A, V)$ additionally incorporating differences in estimated leaf-joint velocities;
(3) $\bm{D}(O, A, V, J_{\text{glo}})$ further including joint position differences from globally reconstructed joints via integrated root velocities; and
(4) $\bm{D}(O, A, V, J_\text{rel})$ using root-relative joint position differences.
We use MPJPE, PA-MPJPE, and Success Rate as evaluation metrics. Quantitative results are shown in Table~\ref{tab:ablation_results}, indicating that adding velocity and root-relative joint position differences improves both kinematic accuracy and stability.

\begin{table}[h]
\centering
\caption{Ablation study on observation configuration.}
\begin{tabular}{lccc}
\toprule
Configuration
& \makecell{MPJPE \\ \textnormal{[mm]} ($\downarrow$)}
& \makecell{PA-MPJPE \\ \textnormal{[mm]} ($\downarrow$)}
& \makecell{Succ.\ Rate \\ \textnormal{[\%]} ($\uparrow$)} \\
\midrule
$\bm{D} (O, A)$ & 290.71 & 51.83 & 90.89 \\
$\bm{D} (O, A, V)$ & 206.50 & 48.97 & 91.25 \\
$\bm{D} (O, A, V, J_{\text{glo}})$ & 187.86 & 48.87 & 93.31 \\
$\bm{D} (O, A, V, J_{\text{rel}})$ & \textbf{182.44} & \textbf{46.47} & \textbf{94.49} \\
\bottomrule
\end{tabular}
\label{tab:ablation_results}
\end{table}
\vspace{-2mm}

%% file: sec/5_coclusion.tex
\section{Conclusion}
\label{sec:conclusion}
We presented GRIP, a physics-based human motion capture framework that integrates four wearable IMUs, insole foot pressure sensors, and physics simulation, and introduced PRISM, a new multimodal dataset capturing diverse human motions and physical interactions. GRIP leverages pressure-based dynamic cues and simulation-based physical constraints to achieve accurate global trajectory estimation and physically plausible pose estimation with a minimal sensor setup. Future directions include designing controllers that can better estimate stable motions under unstable or high-momentum conditions, integrating additional sensing modalities such as cameras or localization sensors to further reduce drift, and extending GRIP to multi-person scenarios and dynamic object interactions for broader applicability in robotics, VR/AR, and biomechanics.

%% file: sec/X_suppl.tex
\clearpage
\setcounter{page}{1}
\maketitlesupplementary

\section{Supplementary Video}

The supplementary video presents an overview of the GRIP method described in Sec.~\ref{sec:method}, including its input data, intermediate network outputs, a visualization of the PRISM dataset, qualitative comparisons with baseline methods (Sec.~\ref{subsec:comparison}), and representative failure cases.

We first visualize the sparse wearable sensor inputs to KinematicsNet and DynamicsNet, including IMU measurements from the wrists and feet and plantar pressure signals from the insole sensors. We then show intermediate outputs of KinematicsNet, including leaf-joint positions, full-body joint positions and joint angles, and key-joint velocities, followed by physically plausible whole-body motion reconstructed by DynamicsNet in the Isaac Gym simulation environment. 
The video also demonstrates the fall-recovery mechanism using a HistoryBuffer that temporarily switches to kinematic predictions after fall detection to safely resume simulation. In addition, we illustrate the sensor configuration used to construct the PRISM dataset and visualize the calibrated and synchronized multimodal sensor data and motion annotations.

Finally, we present qualitative comparisons with baseline methods (PIP~\cite{Yi_PIP}, GlobalPose~\cite{Yi_GlobalPose}, MobilePoser~\cite{Xu_MobilePoser}, and FoRM~\cite{Ying_FoRM}), highlighting differences in global pose accuracy, object interaction, and foot–ground contact naturalness, followed by representative failure cases such as instability during object interaction and challenging motions.

\section{PRISM Dataset Details}
\label{sec:supp_prism}

\subsection{Data Acquisition}
In constructing the PRISM dataset, it is necessary to integrate the various sensor measurements obtained from multiple wearable devices and an optical MoCap system into a unified global coordinate frame and to synchronize all recordings under a consistent temporal reference. This section describes the sensing devices and data acquisition protocol used during collection, the coordinate-frame calibration procedures for each device, and the temporal synchronization process.

\vspace{0.5\baselineskip}
\noindent\textbf{Sensors and Capture Protocol.} 
Data were collected using an optical motion-capture system (OptiTrack), IMU-equipped smartwatches (Apple Watch), IMU-equipped insole-type pressure sensors (Moticon OpenGo), a head-mounted multimodal sensing device (Meta Aria Glass), and IMU devices (Xsens MTw). OptiTrack provides high-precision 3D marker trajectories, and the captured marker data were fitted with Mosh++~\cite{Loper_Mosh} to reconstruct SMPL meshes and high-fidelity body motion. OpenGo includes a 6DoF IMU (accelerometer and gyroscope) and sixteen pressure sensors on each foot, enabling the measurement of vertical forces, pressure distribution, and center of pressure (CoP). The Apple Watch includes a 9DoF IMU (accelerometer, gyroscope, and magnetometer). Aria Glass includes two 9DoF IMUs and simultaneously records forward-facing video and two additional SLAM camera views. Using these videos and IMU data, the Project Aria Machine Perception Service (MPS) was used to obtain environmental point clouds and device trajectories. Furthermore, Xsens MTw units (9DoF IMUs) were attached to the knees and pelvis.

At the beginning of each recording, the subject maintained a T-pose for approximately three seconds while wearing all sensors. This static posture is used to determine the relative orientations of each IMU with respect to the SMPL body model. Starting each capture with the same protocol also reduces ambiguity in the initial pose during real-world operation. After the T-pose, the subject returned to an upright stance and performed three consecutive vertical jumps. The characteristic acceleration peaks observed during the jump motions are later used as robust temporal landmarks for precise synchronization across all sensor streams. The subject then performed the instructed motion sequence.

\vspace{0.5\baselineskip}
\noindent\textbf{IMU Calibration.} 
IMU data were collected from eight body locations using four types of wearable devices: Apple Watch, OpenGo, Aria Glass, and Xsens MTw. Because each device defines its measurements in its own coordinate system, all IMU observations must be transformed into a unified global coordinate frame. We use the coordinate system of the SMPL motion reconstructed from the optical MoCap data using Mosh++ as the global coordinate frame, adopting a z-up convention. At the beginning of each recording session, the subject holds a T-pose facing the positive $y$-axis of the global frame. This protocol enables consistent initialization of all IMU orientations and establishes the reference posture used for the subsequent calibration steps. The following sections describe how raw IMU orientations and accelerations are aligned with the global coordinate frame and further mapped to the corresponding SMPL joint frames.

\begin{figure}[t]
\centering
\includegraphics[width=0.48\textwidth]{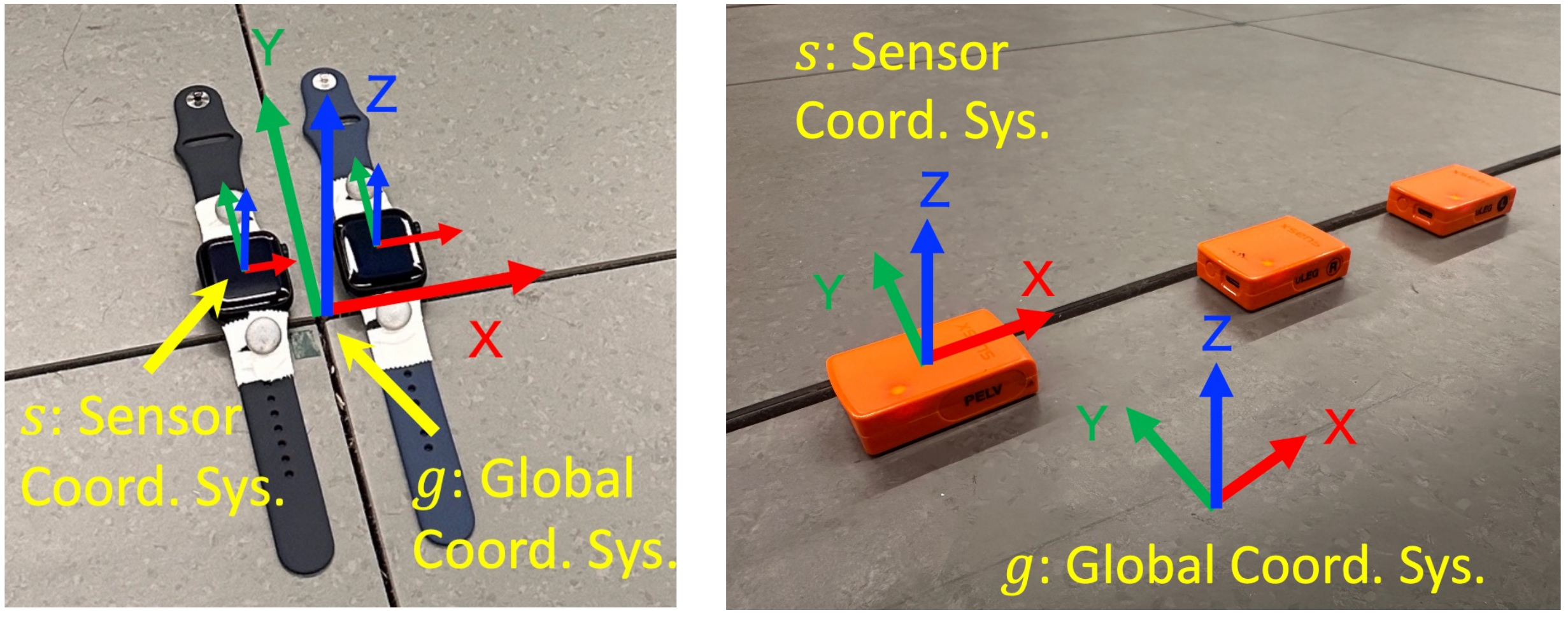}
\caption{Calibration setup for initial orientation. Left: Apple Watch. Right: Xsens MTw sensors.}
\label{fig:watch_xsens}
\vspace{-2mm}
\end{figure}

We first describe the calibration procedure for the Apple Watch and Xsens IMUs, following the notation used in previous works~\cite{Jiang_TIP, lee2024mocapevery}. The relationship between the joint-frame orientation and the raw IMU measurements is expressed as follows:
\begin{equation}
    \bm{R}^{j}_{g} = \bm{R}^{r}_{g} \bm{R}^{s}_{r} \bm{R}^{j}_{s},
\qquad
\bm{a}_{g} = \bm{R}^{r}_{g} \bm{R}^{s}_{r} \bm{a}_{s},
\end{equation}
where $r$ denotes the raw IMU reference frame, and $g$ denotes the global coordinate frame.  
To obtain the joint rotations and accelerations in the global coordinate system at each IMU attachment site, the IMU orientation must be converted into the corresponding joint orientation $\bm{R}^{j}_{g}$, and the raw accelerations defined in the sensor frame must likewise be transformed into the global frame $\bm{a}_{g}$, where $\bm{R}^{s}_{r}$ denotes the raw IMU orientation expressed in the reference sensor frame. Thus, we first estimate the orientation of this reference frame relative to the global frame. Following prior work~\cite{Jiang_TIP, lee2024mocapevery}, before being attached to the user, each IMU sensor is placed on the floor in a specified orientation aligned with the global frame and kept still for 3 seconds (Fig.~\ref{fig:watch_xsens}). By averaging the static IMU signal, we obtain a stable estimate of its orientation at that moment, where $\bm{R}^{s}_{r} = \bm{R}^{g}_{r}$ holds. The inverse $(\bm{R}^{g}_{r})^{-1}$ is then applied to compute $\bm{R}^{r}_{g}$.
In the next step, we compute the joint orientation $\bm{R}^{j}_{g}$ by combining the IMU-to-global rotation with the joint-to-sensor rotation as $\bm{R}^{s}_{g} \bm{R}^{j}_{s}$.  
To determine the joint-to-sensor transform $\bm{R}^{j}_{s}$, the subject takes a T-pose while facing the global $y$-axis. The relationship is given by
\begin{equation}
\bm{R}^{j_T}_{s_T}
= (\bm{R}^{s_T}_{r})^{-1} \bm{R}^{g}_{r} \bm{R}^{j_T}_{g}
\end{equation}
where the subscript $T$ indicates values obtained during the T-pose calibration.  
Assuming the IMU remains rigidly attached to the body during capture, the relative transform $\bm{R}^{j}_{s}$ is constant, allowing us to treat $\bm{R}^{j_T}_{s_T} = \bm{R}^{j}_{s}$.  
Because the joint rotation in the T-pose, $\bm{R}^{j_T}_{g}$, is determined from the SMPL model, the T-pose raw IMU measurements $\bm{R}^{s_T}_{r}$ provide a reliable estimate of $\bm{R}^{j}_{s}$.

\begin{figure}[t]
\centering
\includegraphics[width=0.48\textwidth]{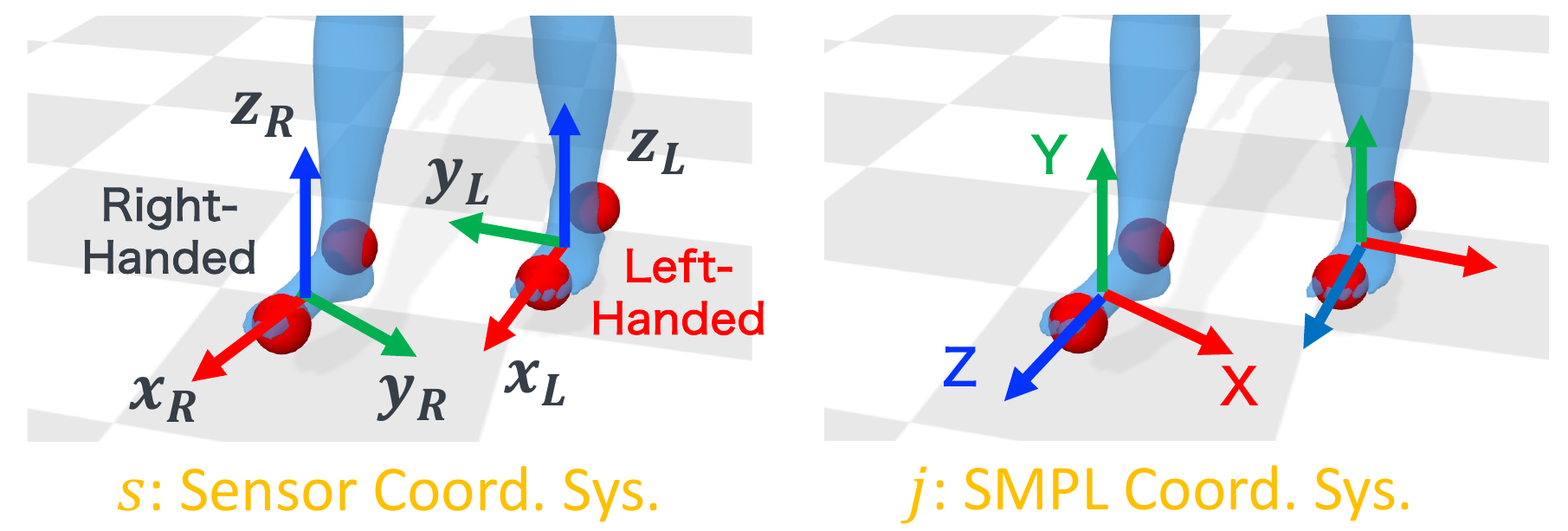}
\caption{Sensor and global coordinate frames for the OpenGo devices, illustrating the handedness difference between sensors.}
\label{fig:opengo}
\vspace{-2mm}
\end{figure}

Next, we describe the calibration procedure for the OpenGo device. Unlike the other IMU devices, the sensor embedded in OpenGo is a 6DoF IMU without a magnetometer, and therefore cannot utilize magnetic-field information for orientation estimation. Following prior work~\cite{Armani2024UIP}, we use the Versatile Quaternion-based Filter (VQF)~\cite{Laidig2023VQF} to estimate the sensor orientation based on the gravity vector. Let the raw sensor coordinate frame be $s$ and the SMPL foot-joint coordinate frame be $j$. Because the left and right OpenGo sensors follow different handedness conventions (Fig.~\ref{fig:opengo}), axis-sign corrections are applied to unify their coordinate systems. The measured angular velocity $\bm{\omega}_{s}$ and acceleration $\bm{a}_{s}$ are mapped into the joint frame using the fixed sensor-to-joint rotation $\bm{R}^{s}_{j}$ as
\begin{equation}
\begin{aligned}
\bm{\omega}_{j,\mathrm{Right}} &= \bm{R}^{s}_{j}\,\bm{\omega}_{s}, \\
\bm{\omega}_{j,\mathrm{Left}}  &= \bm{R}^{s}_{j}\left(\bm{\omega}_{s} \odot [ -1,\,1,\,-1 ]\right), \\
\bm{a}_{j,\mathrm{Right}}      &= \bm{R}^{s}_{j}\,\bm{a}_{s}, \\
\bm{a}_{j,\mathrm{Left}}       &= \bm{R}^{s}_{j}\left(\bm{a}_{s} \odot [ 1,\,-1,\,1 ]\right),
\end{aligned}
\end{equation}
where $\odot$ denotes the element-wise product.
VQF provides the foot orientation in its gravity-aligned frame, denoted as $\bm{R}^j_{\text{vqf}}$. To express this orientation in the global coordinate frame, we align it with the SMPL foot-joint orientation in the T-pose. Let $\bm{R}^{j_T}_{g}$ be the global SMPL foot-joint rotation in the T-pose, and let $\bm{R}^{j_T}_{\text{vqf}}$ be the VQF-estimated sensor orientation at the same moment. The global joint-frame orientation is obtained by directly aligning the VQF-estimated sensor orientation with the calibrated joint orientation in the T-pose:
\begin{equation}
\bm{R}^{j}_{g}
= \bm{R}^{j_T}_{g}\left(\bm{R}^{j_T}_{\text{vqf}}\right)^{-1}\bm{R}^j_{\text{vqf}}.
\end{equation}
The global acceleration is obtained by rotating the raw sensor acceleration using the joint's global orientation:
\begin{equation}
\bm{a}_{g} = \bm{R}^{j}_{g}\,\bm{a}_{j}.
\end{equation}

\begin{figure}[t]
\centering
\includegraphics[width=0.48\textwidth]{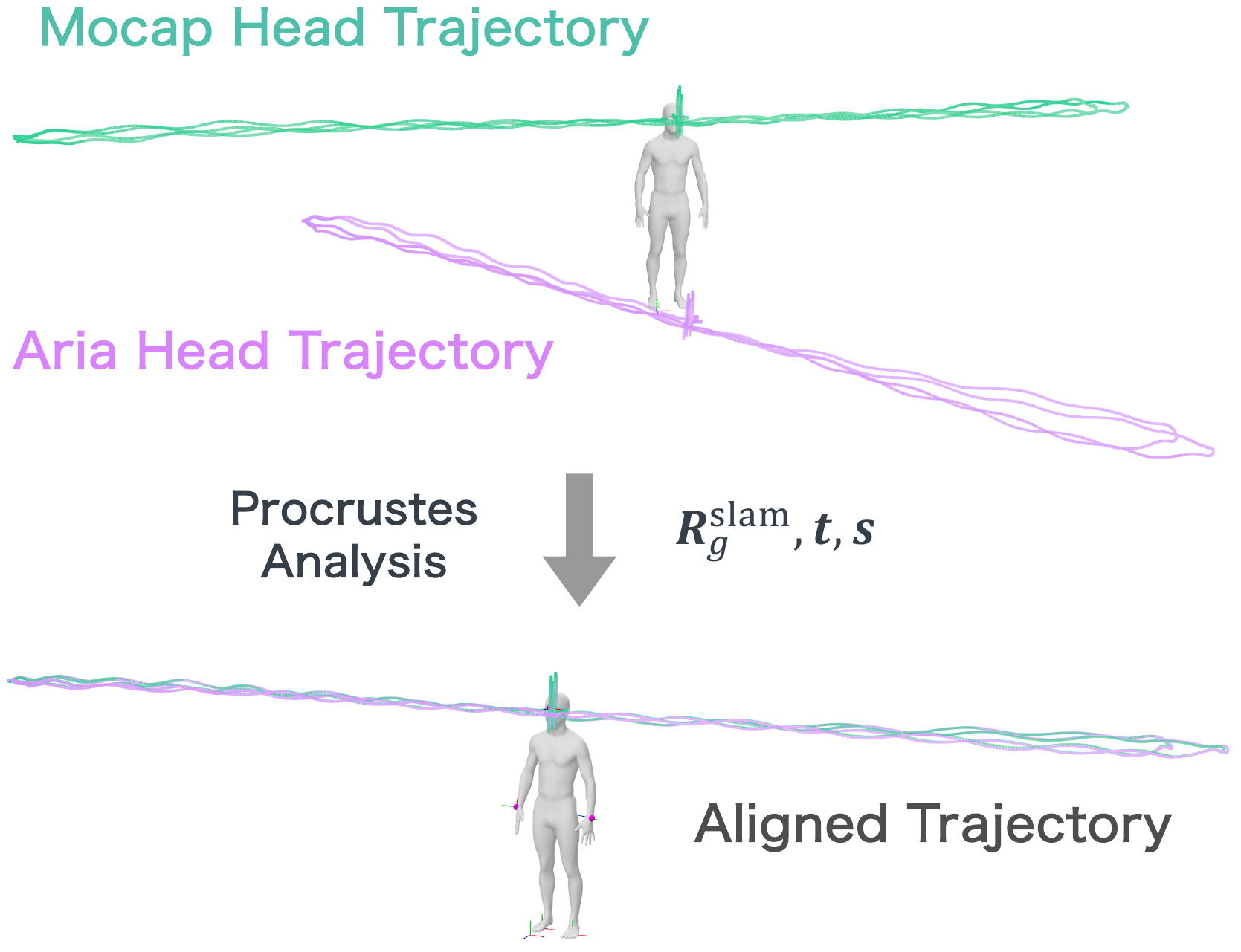}
\caption{Alignment of the Aria Glass SLAM trajectory to the MoCap-based head trajectory using Procrustes analysis.}
\label{fig:aria}
\end{figure}

\begin{figure*}[t!]
\centering
\includegraphics[width=1.0\textwidth]{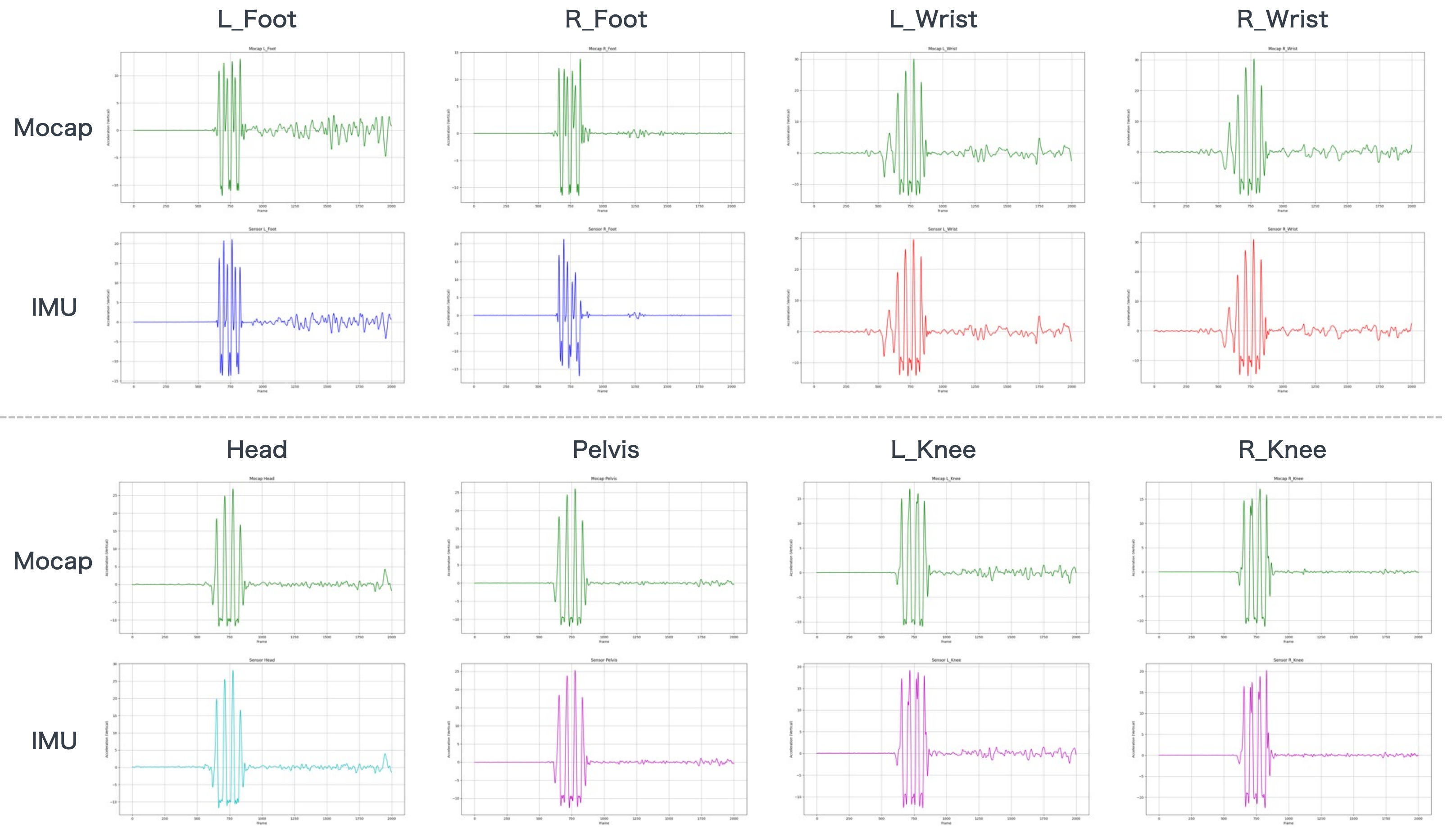}
\caption{Vertical acceleration signals from MoCap data (top row) and from IMU sensors (bottom row) at the same set of body-mounted locations. The characteristic jump-induced peaks provide reliable cues for temporal synchronization across sensor streams.}
\label{fig:synchronization}
\end{figure*}

Finally, we describe the calibration procedure for the Aria Glass.
Aria Glass is equipped with two IMUs (1000\,Hz and 800\,Hz), and in our dataset we use the 1000\,Hz unit, which is downsampled to 100\,Hz to match the other sensors.  
The Aria system defines several coordinate frames, including the IMU frame $i$, the device frame $d$, the SLAM world frame $w$ estimated by MPS, and the Central Pupil Frame (CPF) $c$, which is used to approximate the effective IMU attachment location.  
The global MoCap coordinate system is denoted as $g$.
MPS provides the device orientation as $\bm{R}^{d}_{w}$ in the SLAM coordinate system, and factory calibration further provides the fixed extrinsic rotations $\bm{R}^{i}_{d}$ (IMU to device) and $\bm{R}^{c}_{i}$ (IMU to CPF).  
The transformation from the SLAM frame to the global MoCap frame, $\bm{R}^{w}_{g}$, is estimated by leveraging the high accuracy of the SLAM trajectory produced by MPS.  
Specifically, as illustrated in Fig.~\ref{fig:aria}, we align the Aria Glass trajectory to the MoCap trajectory of the SMPL-based CPF (central pupil) position across the entire sequence using Procrustes analysis, which yields the rotation $\bm{R}^{w}_{g}$, translation $\bm{t}$, and scale $s$.
Using these transformations, the CPF orientation and IMU accelerations can be expressed in the global coordinate system as
\begin{equation}
\begin{aligned}
\bm{R}^{c}_{g} &= \bm{R}^{w}_{g}\,\bm{R}^{d}_{w}\,\bm{R}^{i}_{d}\,\bm{R}^{c}_{i}, \\
\bm{a}_{g}     &= \bm{R}^{w}_{g}\,\bm{R}^{d}_{w}\,\bm{R}^{i}_{d}\,\bm{a}_{i},
\end{aligned}
\end{equation}
where $\bm{a}_{i}$ denotes the raw acceleration measured in the IMU frame.  
These quantities are used as the head IMU’s global orientation and global acceleration in the dataset.

\begin{table*}[t]
\centering
\caption{Comparison of human motion datasets utilizing insole-type foot pressure sensors.}
\begin{tabular}{lcccccc}
\toprule
Dataset & Year & Foot IMUs & Other IMUs & Env. Models & Motion Diversity & MoCap Method \\
\midrule
PSU-TMM100~\cite{Scott_Stability} & 2020 & No & No & No & Low (Tai Chi) & Optical \\
UnderPressure~\cite{Mourot_UnderPressure} & 2022 & Yes (2) & No & No & Medium (Locomotion) & Inertial \\
MMVP~\cite{Zhang_MMVP} & 2024 & No & No & No & Medium (Exercise) & RGBD \\
HAMPI~\cite{Ying_FoRM} & 2025 & Yes (4) & No & No & High (Diverse Action) & Inertial + RGB \\
\textbf{PRISM (Ours)} & 2025 & Yes (2) & Yes (6) & Yes & High (Diverse Action) & Optical \\
\bottomrule
\end{tabular}
\label{tab:dataset_comp}
\end{table*}

\vspace{0.5\baselineskip}
\noindent\textbf{Time Synchronization.}
All sensors used in this study (OptiTrack, Apple Watch, OpenGo, Aria Glass, and Xsens MTw) record data at 100 Hz. Since the measurements on each device are started manually, their recording start times are not aligned. Therefore, a synchronization procedure is required to align all sensor streams to a common temporal reference. In our dataset, we use the vertical acceleration peaks generated during the three vertical jumps performed immediately after the T-pose as temporal landmarks for synchronization. As described in the previous section, all IMUs are transformed into a unified global coordinate system through calibration. This allows the vertical acceleration components derived from each IMU’s measurements to be compared directly.
Meanwhile, the SMPL mesh obtained from MoCap provides only positional information; therefore, the corresponding reference acceleration is computed by applying finite differences to the time-series vertex positions at each IMU attachment site, converting positions to velocities and subsequently to accelerations.

For synchronization, the vertical acceleration waveform of each IMU is slid over the reference SMPL-derived vertical acceleration waveform using a sliding-window approach, and similarity (cross-correlation) is evaluated at each offset. The temporal offset that achieves the highest similarity is taken as the time shift for that IMU stream, aligning all sensors to a common time axis. Fig.~\ref{fig:synchronization} shows an example of the vertical acceleration signals computed from MoCap-based SMPL at the eight IMU attachment locations (wrists, feet, knees, waist, and head), alongside those from the corresponding IMU sensors, illustrating how the characteristic peaks from the three jumps serve as reliable synchronization cues.

\subsection{Dataset Composition.}
The PRISM dataset consists of 1,275 sequences divided into 1,021 for training and 254 for testing, each 10~seconds long (1,000~frames). Data were collected from six subjects (four male and two female), totaling approximately 12,750~seconds ($\approx$~3.5~hours) of motion. Each sequence includes synchronized multimodal signals, such as MoCap-based reference poses, IMU and insole pressure data, environmental point clouds, camera trajectories, and object models for interaction.

The dataset covers a wide range of motion categories.
First, Basic Locomotion includes walking along predefined paths and random trajectories (Walking Square / Walking Random), as well as jogging along predefined or random paths (Jogging Square / Jogging Random).
Next, Slow Movements consist of stretching, two-foot and one-foot balancing, squatting, and lunge walking.
Furthermore, Fast Sports Actions include whole-body athletic movements such as sidestepping, forward jumping, golf, baseball, tennis, soccer, and direction-switching movements.
Finally, Interaction Behaviors involve movements that include physical interaction with objects, such as stepping onto boxes or stairs (Stepping Boxes / Stepping Stair), sitting on boxes (Sitting Boxes), and object carrying.

\subsection{Comparison with Existing Datasets.}

In recent years, several foot-sensing motion datasets leveraging insole pressure sensors have been released, contributing significantly to the understanding of human motion (Table~\ref{tab:dataset_comp}). However, existing datasets exhibit inherent trade-offs across aspects such as IMU availability, environmental information, motion diversity, and annotation accuracy. As a result, none of them is sufficient for comprehensively modeling physically consistent full-body motion together with interactions involving the environment and objects.

PSU-TMM100~\cite{Scott_Stability} is a dataset that combines insole-type pressure sensors with optical motion capture, consisting of Tai Chi movements. While it provides plantar pressure distributions, it does not include any IMU measurements. UnderPressure~\cite{Mourot_UnderPressure} contains long recordings of basic locomotion such as walking, jogging, and skipping, and is characterized by the inclusion of both plantar pressure sensors and foot IMUs. It also includes a limited set of environment-interaction behaviors, such as stair ascent/descent and stepping onto objects, but neither environment geometry nor object models are provided. MMVP~\cite{Zhang_MMVP} records RGB-D video of body movements together with plantar pressure measurements, and features several fast exercise motions, but does not provide IMU data. HAMPI~\cite{Ying_FoRM}, the most recent foot-sensing dataset, includes plantar pressure, foot IMUs, and SMPL motions generated by combining inertial motion capture with monocular video-based pose estimation. Although it covers a wide variety of actions, the IMUs are restricted to the feet, making the dataset unsuitable for studies that require wrist or full-body IMUs. Moreover, some motion-capture errors remain in the form of unnatural body tilts or elbow twisting, making the annotations not fully reliable for physics-based approaches such as GRIP.

Taken together, existing datasets do not simultaneously provide (i) detailed foot-sensing measurements, (ii) full-body inertial information, (iii) environment interaction data, (iv) a broad range of motions, and (v) accurate full-body mesh annotations within a single framework. PRISM addresses these gaps by combining SMPL motion reconstructed from optical MoCap with eight IMUs placed at common attachment sites (knees, pelvis, wrists, and head) in addition to the feet, along with insole-type pressure sensors. PRISM also includes environmental point clouds and object models, and covers a wide variety of motions ranging from basic locomotion to slow movements, fast sports activities, and interactions with objects. Through this combination, PRISM serves as a unified dataset suitable for integrated analysis of full-body motion and foot–ground interaction dynamics in everyday scenarios, supporting research that requires physically consistent motion estimation.

\section{Implementation Details}
\noindent\textbf{Reward Design in DynamicsNet.}
The reward function follows the Perpetual Humanoid Control (PHC) framework~\cite{Luo_PHC},  
and consists of three components: an Adversarial Motion Prior (AMP)~\cite{Peng_AMP}, an imitation reward,  
and an energy penalty~\cite{Fu2022DeepWholeBodyControl}.  
The total reward is formulated as:
\begin{equation}
r_t = 0.5r_t^{\text{amp}} + 0.5r_t^{\text{imit}} + r_t^{\text{energy}},
\end{equation}
where $r_t$ denotes the total reward at timestep $t$.

The AMP reward $r_t^{\text{amp}}$ encourages realistic and natural motions by training a discriminator $D$ to distinguish generated motions from real human motion.
The discriminator takes as input the self-observation $\bm{O}_t^{\text{self}}$, which consists of a temporally ordered sequence of motion features over $W$ consecutive frames, maintained as a sliding window that is updated at every timestep.
The AMP reward is computed as:
\begin{equation}
r_t^{\text{amp}} = -\log\left(1 - \sigma\big(D(\bm{O}_t^{\text{self}})\big)\right),
\end{equation}
where $\sigma(\cdot)$ denotes the sigmoid function and $D(\bm{O}_t^{\text{self}})$ represents the discriminator logit output for the self-observation $\bm{O}_t^{\text{self}}$.
This reward formulation encourages the policy to generate motions that the discriminator classifies as real, i.e.,
$\sigma\big(D(\bm{O}_t^{\text{self}})\big) \to 1$,
thereby maximizing $r_t^{\text{amp}}$.
The discriminator is trained using a standard Generative Adversarial Network (GAN) objective with gradient penalty regularization:
\begin{equation}
\begin{aligned}
\mathcal{L}_D
&= \mathbb{E}_{\bm{O}^{\text{self}}_\pi}
\left[\log\left(1 - \sigma\big(D(\bm{O}^{\text{self}}_\pi)\big)\right)\right] \\
&\quad + \mathbb{E}_{\bm{O}^{\text{self}}_{\text{ref}}}
\left[\log \sigma\big(D(\bm{O}^{\text{self}}_{\text{ref}})\big)\right] \\
&\quad + \lambda_{\text{gp}}
\mathbb{E}_{\bm{O}^{\text{self}}_{\text{ref}}}
\left[\left\|\nabla_{\bm{O}} D(\bm{O}^{\text{self}}_{\text{ref}})\right\|^2\right],
\end{aligned}
\end{equation}
where $\bm{O}^{\text{self}}_\pi$ denotes self-observations generated by the policy,
$\bm{O}^{\text{self}}_{\text{ref}}$ denotes self-observations from the reference motion dataset,
$\lambda_{\text{gp}}$ is the gradient penalty coefficient,
and $\nabla_{\bm{O}} D(\cdot)$ represents the gradient of the discriminator output with respect to the input observation.

The imitation reward measures the consistency between the simulated and reference motion states.  
It is computed based on the differences in joint positions, rotations, linear velocities, and angular velocities:
\begin{equation}
\begin{split}
r_t^{\text{imit}} ={}&
w_p e^{-100\|\hat{\bm{p}}_t - \bm{p}_t\|} +
w_\theta e^{-100\|\hat{\bm{\theta}}_t \ominus \bm{\theta}_t\|} \\
&+
w_v e^{-10\|\hat{\bm{v}}_t - \bm{v}_t\|} +
w_\omega e^{-0.1\|\hat{\bm{\omega}}_t - \bm{\omega}_t\|},
\end{split}
\end{equation}
where $w_p$, $w_\theta$, $w_v$, and $w_\omega$ are weighting coefficients for each term,  
and $\ominus$ denotes the rotation difference between the reference and simulated orientations. 
The hatted quantities (e.g., $\hat{\bm{p}}_t$, $\hat{\bm{\theta}}_t$) represent reference motion values,  
while the unaccented ones correspond to simulated states.

In addition, the energy penalty $r_t^{\text{energy}}$ is computed as the negative mechanical power of the humanoid, obtained from the element-wise product of joint torques $\boldsymbol{\tau}_t$ and joint angular velocities $\boldsymbol{\omega}_t$. Specifically, the penalty is defined as
\begin{equation}
r_t^{\text{energy}} = -\alpha \, |\boldsymbol{\tau}_t \odot \boldsymbol{\omega}_t|,
\end{equation}
where $\alpha$ is a power coefficient (default 0.0005). This term encourages energy-efficient control while excluding the first three frames to avoid initialization artifacts.

\begin{figure*}[t]
\centering
\includegraphics[width=1.0\textwidth]{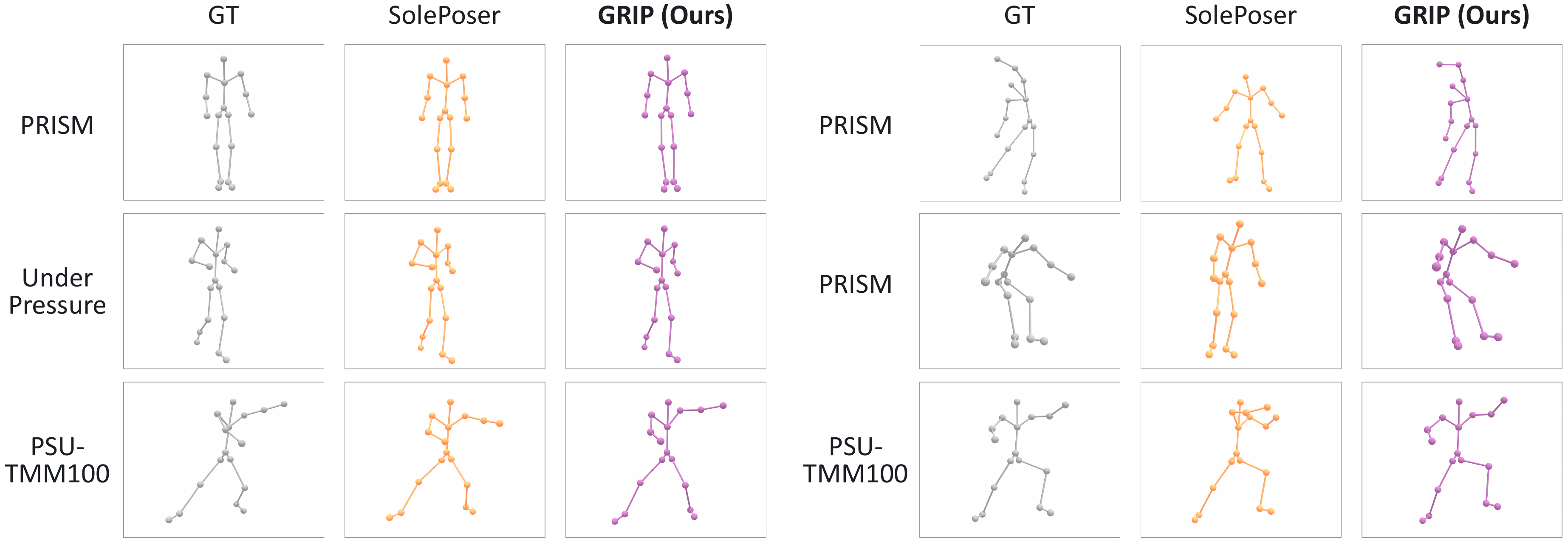}
\caption{Comparison between SolePoser and our method using the same 17 joint locations extracted from the SMPL model.}
\label{fig:soleposer}
\end{figure*}

\vspace{0.5\baselineskip}
\noindent\textbf{Networks and Training.}
The proposed framework consists of two stages: KinematicsNet, which estimates kinematic states, and DynamicsNet, which generates physically consistent motion within a physics simulator.  
Training is performed in two stages: first, KinematicsNet is trained independently, and then DynamicsNet is trained within a physics simulation environment. During the training of DynamicsNet, the parameters of KinematicsNet are frozen and used as a motion-state estimator.

KinematicsNet comprises four submodules: LP (leaf-joint position), FP (full-joint position), LV (leaf-joint velocity), and FA (full-joint angle).  
Each submodule is first trained independently to ensure stable estimation, and then all modules are jointly trained in an integrated manner.  
Following previous works~\cite{Yi_PIP, zhang_DynaIP, Yi_PNP, Yi_GlobalPose}, we initialize the hidden states of each submodule RNN using the ground-truth data of the first frame, to avoid ambiguity caused by uncertain initial poses. This initialization is applied only to the first frame of each sequence.
Training was conducted for 2000 epochs using the Adam optimizer (learning rate $1\times10^{-5}$, batch size 256). Following TIP~\cite{Jiang_TIP}, IMU signals were processed with a low-pass filter to remove high-frequency noise. All training was performed on data sampled at 100~Hz.

DynamicsNet is a policy network trained within a physics simulation environment to control a torque-driven humanoid model.  
The policy network is implemented as a multilayer perceptron (MLP) consisting of six layers with sizes [2048, 1536, 1024, 1024, 512, 512] and SiLU activations. The network outputs target joint angles, which are converted into joint torques through PD control.  
In the early training stage, an early termination condition is applied: an episode is terminated when any joint position error exceeds a threshold of $\tau_e = 0.25$~m, which accelerates convergence and stabilizes the learning of basic motion patterns. We train DynamicsNet for 500 epochs across 10 iterations under this early-termination regime using 4096 parallel environments, a policy minibatch size of 32{,}768, and an AMP minibatch size of 8192.  
Afterward, the early termination constraint is removed to enable training over long motion sequences, and an additional 500 epochs across 20 iterations are performed under the full-length setting. 
In addition, to enable fall-recovery mechanism during inference and allow the simulation to continue even after the humanoid falls, we introduce a fall-detection condition in which the humanoid is considered fallen when the root joint height drops below 0.30 m and the AMP discriminator probability $\rho_t$ falls below 0.7.

Training takes about 24 hours for KinematicsNet and 48 hours for DynamicsNet.
During inference, KinematicsNet and DynamicsNet require approximately 1.5 ms and 26.5 ms per frame, respectively.
The sequential prediction in KinematicsNet is designed to improve accuracy, with limited computational overhead.

\vspace{0.5\baselineskip}
\noindent\textbf{Computational Environment.}
Experiments were conducted on Ubuntu 20.04.6 LTS.
The system is equipped with an AMD EPYC~7543 processor configuration 
providing 64~cores (128~threads), 251~GB RAM, and eight NVIDIA RTX~A6000 GPUs 
with 48~GB VRAM each. 
All neural networks were implemented in PyTorch and optimized using the Adam optimizer.
Physics simulations and reinforcement learning were performed using Isaac Gym 
in a CUDA~11.8--enabled environment with NVIDIA driver version~570.133.07.

\section{Additional Experimental Results}
\noindent\textbf{Qualitative Comparison with SolePoser. }
In this section, we present a qualitative comparison between SolePoser~\cite{Wu_SolePoser} and our proposed method, GRIP, which could not be included in the main paper due to space limitations. SolePoser estimates root-relative full-body pose using only the foot-mounted IMUs and foot pressure data obtained from insole sensors. Fig.~\ref{fig:soleposer} shows results on the PRISM, UnderPressure, and PSU-TMM100 datasets.

SolePoser produces reasonable estimates for relatively simple poses and periodic locomotion such as walking and jogging. However, due to the limitation of using only foot sensors, its accuracy degrades significantly for motions involving substantial upper-body movement or complex postural changes, and several failure cases can be observed. In contrast, GRIP incorporates wrist-mounted IMUs in addition to foot IMUs and plantar-pressure sensing, allowing it to capture a broader range of whole-body inertial information. As a result, our method reconstructs full-body poses more accurately even in non-periodic and complex motions where SolePoser struggles.
These observations indicate that GRIP can estimate whole-body dynamics that cannot be recovered from foot-only sensing, by combining IMUs at the extremities (hands and feet) with foot pressure information.